\documentclass[10pt,twocolumn,letterpaper]{article}

\PassOptionsToPackage{table}{xcolor}
\usepackage{iccv}
\usepackage[accsupp]{axessibility}
\usepackage{amsmath,amsfonts}

\usepackage{amssymb}
\usepackage{algorithmic}
\usepackage{algorithm}
\usepackage{array}
\usepackage{textcomp}
\usepackage{stfloats}
\usepackage{float}
\usepackage{url}
\usepackage{verbatim}
\usepackage{graphicx}
\usepackage{booktabs}
\usepackage{bm}
\usepackage{pifont}
\usepackage{enumitem}
\usepackage{soul}
\usepackage{makecell}
\usepackage{tabularx}
\usepackage{multirow}
\usepackage{siunitx}
\usepackage{balance}
\usepackage{lscape}
\usepackage[utf8]{inputenc}
\usepackage[T1]{fontenc}
\usepackage{pmboxdraw} 
\usepackage{tocloft}
\definecolor{cvprblue}{rgb}{0.21,0.49,0.74}
\usepackage{listings}
\usepackage{rotating}
\usepackage{makecell}
\usepackage{caption}
\usepackage{multicol}
\usepackage{wrapfig}
\usepackage{colortbl}
\usepackage{arydshln}

\usepackage{xspace}
\usepackage{rotating}
\usepackage{framed}
\usepackage{bbm}
\usepackage{gensymb}
\usepackage[export]{adjustbox}
\definecolor{iccvblue}{rgb}{0.21,0.49,0.74}
\usepackage[pagebackref,breaklinks,colorlinks,allcolors=iccvblue]{hyperref}
\usepackage[capitalize]{cleveref}

\definecolor{vscode-light-background}{RGB}{240,239,239}  
\definecolor{vscode-foreground}{RGB}{0,0,0}               
\definecolor{vscode-blue}{RGB}{56,92,170}                 
\definecolor{vscode-green}{RGB}{68,191,189}                
\definecolor{vscode-red}{RGB}{237,90,38}                  
\definecolor{vscode-purple}{RGB}{134,0,179}                
\definecolor{vscode-orange}{RGB}{214,103,0}                

\crefname{section}{Sec.}{Secs.}
\Crefname{section}{Section}{Sections}
\Crefname{table}{Table}{Tables}
\crefname{table}{Tab.}{Tabs.}

\def \ie {\emph{i.e.},}
\def \eg {\emph{e.g.},}

\newcolumntype{Y}{>{\centering\arraybackslash}X}

\definecolor{iccvblue}{rgb}{0.21,0.49,0.74}
\usepackage[pagebackref,breaklinks,colorlinks,allcolors=iccvblue]{hyperref}
\usepackage{multirow}
\usepackage{graphicx}

\def\paperID{11} 
\def\confName{ICCVW}
\def\confYear{2025}

\title{Dual Orthogonal Guidance \\ for Robust Diffusion-based Handwritten Text Generation
}

\author{
Konstantina Nikolaidou$^{1}$\thanks{equal contribution} 
\quad
George Retsinas$^{2}$\footnotemark[1]
\quad Giorgos Sfikas$^{3}$ \\ \quad Silvia Cascianelli$^4$ \quad Rita Cucchiara$^4$ \quad Marcus Liwicki$^1$ \\ \\ 
\begin{tabular}{ccc}
\makecell{$^1$Luleå University of Technology}&
\makecell{$^2$National Technical University of Athens} \\
\makecell{$^3$University of West Attica} &
\makecell{$^4$University of Modena and Reggio Emilia} \\
\end{tabular}
}

\begin{document}
\maketitle
\begin{abstract}

Diffusion-based Handwritten Text Generation (HTG) approaches achieve impressive results on frequent, in-vocabulary words observed at training time and on regular styles. 
However, they are prone to memorizing training samples and often struggle with style variability and generation clarity. 
In particular, standard diffusion models tend to produce artifacts or distortions that negatively affect the readability of the generated text, especially when the style is hard to produce. 
To tackle these issues, we propose a novel sampling guidance strategy, Dual Orthogonal Guidance (DOG), that leverages an orthogonal projection of a negatively perturbed prompt onto the original positive prompt. 
This approach helps steer the generation away from artifacts while maintaining the intended content, and encourages more diverse, yet plausible, outputs.
Unlike standard Classifier-Free Guidance (CFG), which relies on unconditional predictions and produces noise at high guidance scales, DOG introduces a more stable, disentangled direction in the latent space. 
To control the strength of the guidance across the denoising process, we apply a triangular schedule: weak at the start and end of denoising, when the process is most sensitive, and strongest in the middle steps.
Experimental results on the state-of-the-art DiffusionPen and One-DM demonstrate that DOG improves both content clarity and style variability, even for out-of-vocabulary words and challenging writing styles.
\end{abstract}    
\section{Introduction}
\label{sec:intro}

Handwritten Text Generation (HTG), or Styled HTG, is a task that has only relatively recently gained traction compared to the more ``traditional'' Document Imaging tasks, like Handwritten Text Recognition (HTR)~\cite{kang2022pay,cascianelli2022boosting,retsinas2022best,retsinas2021seq2seq,retsinas2024enhancing} or Keyword Spotting (KWS)~\cite{retsinas2016keyword,retsinas2018efficient,jemni2025st}, in terms of effective models and methods.
One motivation for HTG systems is user personalization in digital applications, where it can be useful in aiding individuals with physical impairments to produce handwritten notes in a personalized style.
Another crucial goal for HTG is to play the role of an efficient tool for data augmentation, acting in an auxiliary way for other main, downstream Document Analysis tasks~\cite{diaz2025survey}. 
This is especially useful in low-resource contexts~\cite{nikolaidou2022survey},
where inadequately documented scripts or languages with few writers or with few digitized, annotated training samples constitute a serious impediment to creating automated document-imaging tools.

\begin{figure}[t]
  \centering
  \includegraphics[width=\columnwidth]{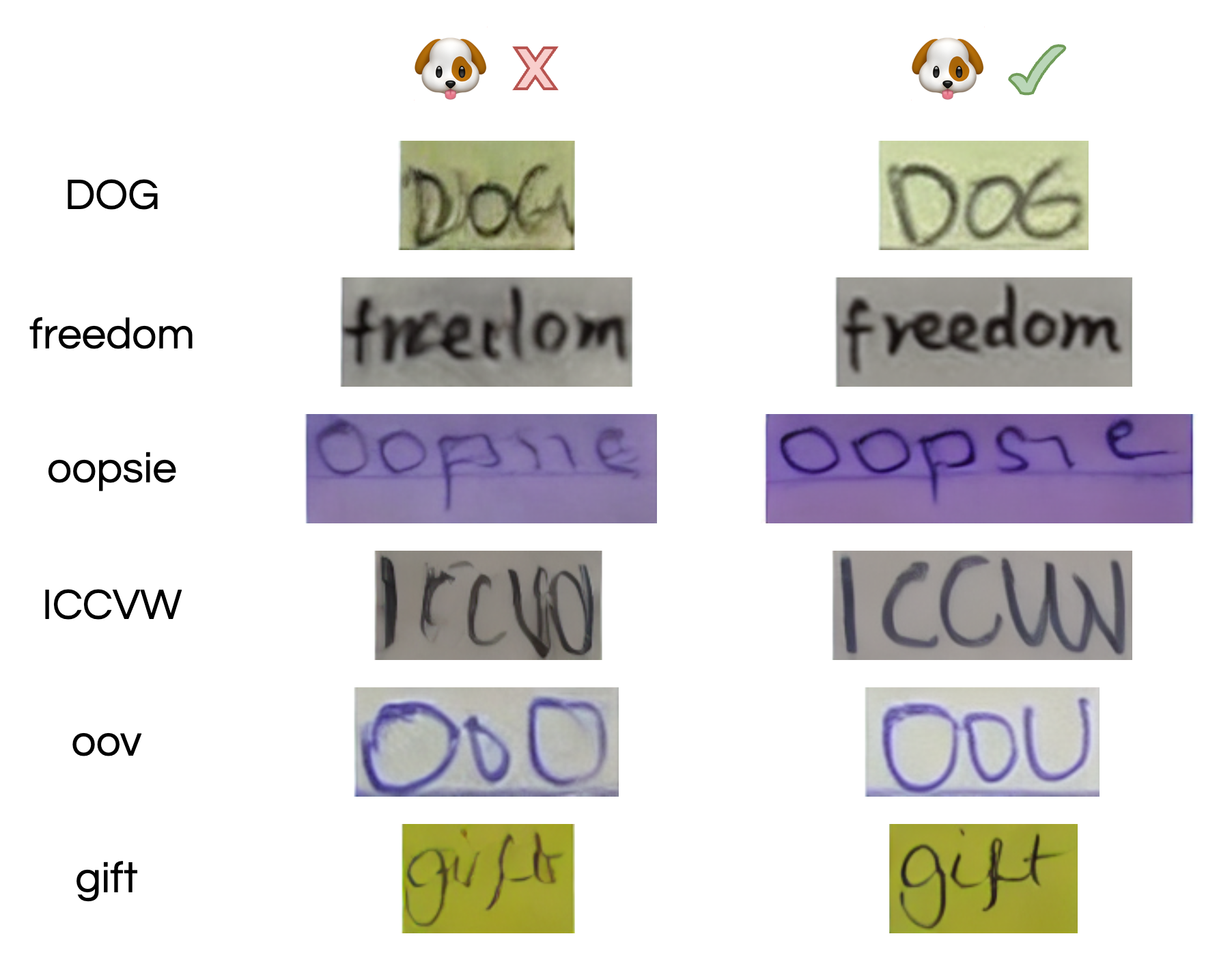}
  \caption{Qualitative examples of generation without (left) and with (right) our proposed \textit{DOG} guidance strategy.}
  \label{fig:teaser}
  \vspace{-12pt}
\end{figure}

Diffusion-based HTG models have recently shown promising results in producing readable handwritten images replicating an existing style. 
However, two major issues persist:

\begin{enumerate}
    \item They frequently generate artifacts that degrade content clarity, even for common, in-distribution examples.
    \item They tend to memorize style-content pairs seen during training, making them brittle in low-data regimes or for unseen combinations.
\end{enumerate}

To partially address the first issue, Classifier-Free Guidance (CFG)~\cite{ho2021classifierfree} has been proposed as a sampling method that interpolates between unconditional and conditional predictions, typically improving alignment with the target condition. However, CFG introduces a trade-off: high guidance scales may help with clarity but often result in over-saturation or degraded detail.

In this work, we introduce Dual Orthogonal Guidance (DOG), a guidance strategy that pushes generation along an orthogonal direction derived from a negatively perturbed version of the conditioning prompt. 
The idea is to encourage clearer generation by suppressing entangled distortions from a perturbed condition (see Fig.~\ref{fig:teaser}), while still enabling structured variation.
Unlike CFG, which relies on unconditional sampling, DOG uses a negative prompt derived from the actual condition (\eg~noised style or content), which is then orthogonalized with respect to the original. 
This produces a more targeted and stable modification of the sampling trajectory. 
To control the influence of this guidance throughout the diffusion process, we use a triangular schedule, which limits the effect at early and late steps, where the process is most sensitive, and maximizes it in the middle steps, where it can influence the structure without causing instability. 


In particular, our main contributions are the following:
\begin{itemize}
    \item \emph{Dual Orthogonal Guidance (DOG):} 
    A test-time sampling strategy that introduces an orthogonal direction derived from a negative prompt, enabling clearer generation and controlled variability.
    \item \emph{Stability through triangular scheduling:} We modulate the guidance scale across timesteps, peaking mid-process to avoid distorting global structure or introducing artifacts.
    \item \emph{Plug-and-play integration:} DOG is model-agnostic, requires no retraining, and can be applied directly to pre-trained diffusion-based HTG models.
     \item \emph{Compelling Performance:} We beat the state-of-the-art in terms of posterior sampling variety, with tangible benefits in terms of qualitative and quantitative results.
\end{itemize}

The remainder of this paper is structured as follows.
In Section \ref{sec:related_work}, we review the related work.
In Section \ref{sec:method}, we describe the proposed Dual Orthogonal Guidance.
Section \ref{sec:experiments} presents numerical experiments and showcases illustrative qualitative results.
Finally, we conclude the paper with Section \ref{sec:conclusion}, where we summarize our contributions and discuss future work.

\section{Related work}
\label{sec:related_work}

\vspace{1.1mm}
\noindent
\textbf{Styled HTG \& Classifier-free Guidance}.
\textit{Styled HTG} is the task of generating handwritten images given a style and a content condition.
Early HTG methods relied on extensive hand-crafted feature extraction and text resynthesizing using 
rule-based methods~\cite{wang2004image,lin2007style,thomas2009synthetic}. 
Later, recurrent architectures demonstrated the ability to generate handwriting \cite{graves2013generating}. 
Building on early methods, GAN-based techniques~\cite{alonso2019adversarial,Fogel2020ScrabbleGANSV,Davis2020TextAS,kang2020ganwriting,kang2021content,mattick2021smartpatch,gan2022higan+}, including several that integrate transformer architectures~\cite{Bhunia_2021_ICCV,Pippi2023HandwrittenTG,vanherle2024vatr++}, have enhanced handwriting generation. 
However, these approaches still face challenges such as training instability and limited diversity \cite{farnia2020gans,arjovsky2017towards,mescheder2018training}.
More recently, Diffusion Models \cite{zhu2023conditional,nikolaidou2023wordstylist,dai2025one,nikolaidou2024diffusionpen,brandenbusch2024semi,gurav2025word} have been put forward as an alternative to GAN-based models.
While the paradigm of diffusion forms a cohesive foundation for styled HTG, several problems have proved to be non-straightforward to solve in a satisfactory manner, including style variability, rare character combinations, and training data memorization.
\textit{Classifier-free guidance (CFG)}~\cite{ho2021classifierfree} has shown improvement in the quality of the generation by performing a linear interpolation between the conditional and unconditional estimations,
$\lambda \epsilon(x|c) + (1-\lambda)\epsilon(x|\emptyset)$.
A few works have showcased the effect of CFG in HTG by exploring the standard approach~\cite{dai2025one,ding2023improving,mayr2024zero,brandenbusch2024semi}.
However, only~\cite{brandenbusch2024semi} explores the effect of CFG on the generation.

\vspace{1.1mm}
\noindent
\textbf{Negative Prompting in Diffusion Models}.
Diffusion Models are state-of-the-art latent variable generative models that are consistently setting new benchmarks in numerous and diverse tasks \cite{sohl2015deep}, including those that pertain to one or another form of prompting, in the sense of a textual condition to the model \cite{ho2020denoising,croitoru2023diffusion}.
\textit{Negative Prompting} seeks to steer the guidance away from unwanted attributes. 
In \cite{liu2022compositional},
an objective that may include multiple conditions is broken down to a sum of composing directions in the latent space.
Unwanted conditions are then assigned a negative weight, in principle allowing the end-user to specify the set of conditions at will. 
This translates to a simple but very effective test-time algorithm.
Perp‐Neg~\cite{armandpour2023re} is another sampling algorithm for standard Text-to-Image Diffusion Models, which computes a negative gradient that is perpendicular to the main prompt. 
The rationale related to this choice is that two conditions should not be taken \emph{a priori} to be conditionally independent; moving in the orthogonal direction ensures that unwanted details are suppressed without interfering with the primary semantic content.
Both works underscore the importance of carefully adjusting the positive and negative cues to enhance the fidelity and controllability of the generated images.
Our proposal is inspired by this previous work, and puts forward a technique that is also based on test-time guidance of sampling.




Interestingly, a very recent and independent work also explores the use of orthogonal projections in diffusion guidance~\cite{sadat2024eliminating}.
This method, named Adaptive Projected Guidance (APG), applies a perpendicular projection of the unconditional CFG term to improve saturation in text-to-image generation for the case of high guidance scales. 
While our approach was developed independently and is specifically tailored to HTG, both our work and~\cite{sadat2024eliminating} 
highlight the benefit of introducing orthogonal components to steer the generative process without conflicting with the main conditioning. 
However, unlike DOG, APG further assigns a small weight to the parallel component and lacks a scheduling mechanism to modulate the influence of guidance across denoising steps. 
As we show, the absence of such scheduling can make guidance scale selection delicate, introducing a trade-off between control and artifact-free generation. 
In our work, we compare both existing guidance strategies with our proposed DOG. 

\section{Dual Orthogonal Guidance (DOG)}
\label{sec:method}

\begin{figure*}[ht]
  \centering
  \includegraphics[width=\textwidth]{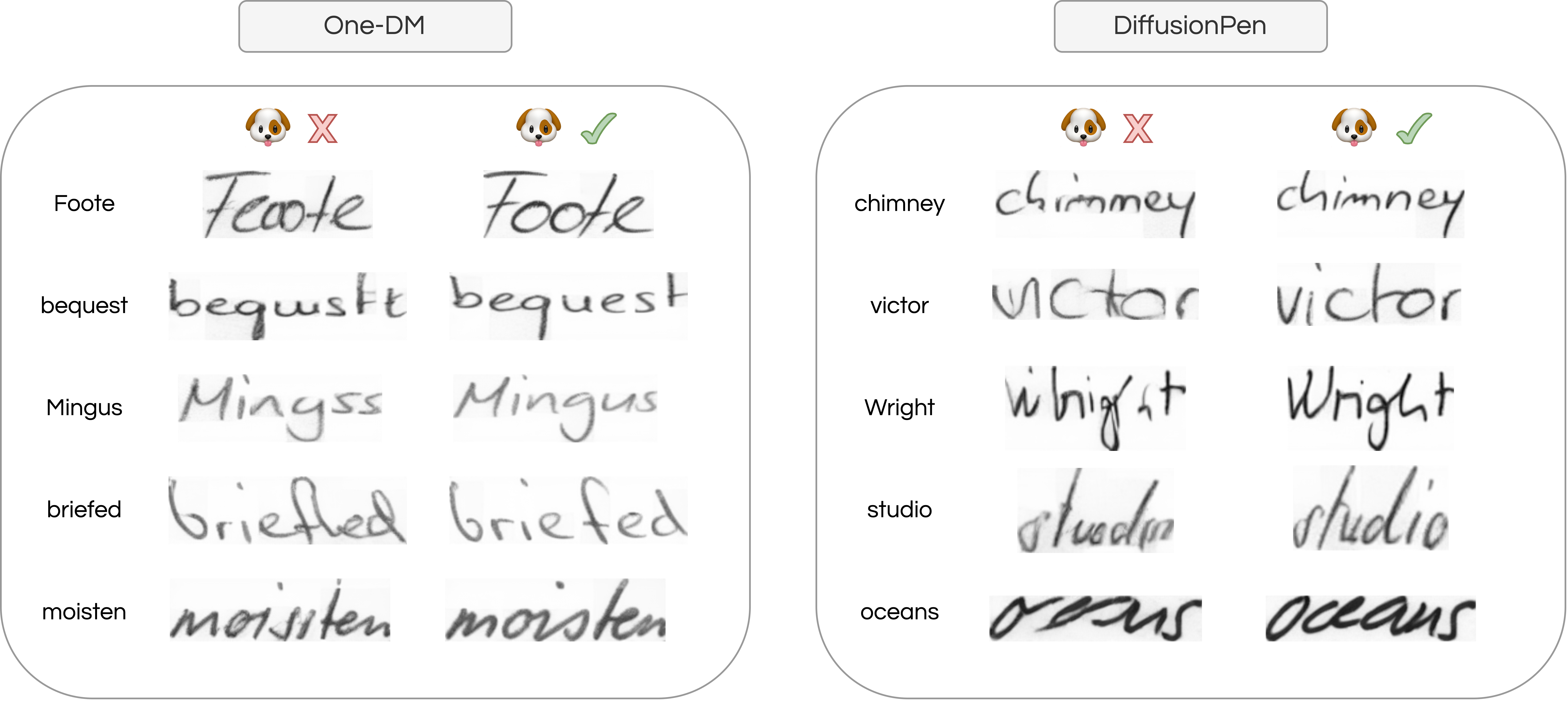}
  \caption{
  Qualitative results of the application of the proposed DOG guidance strategy at inference time to the Diffusion-based OneDM and DiffusionPen HTG approaches.}
  \label{fig:oov_onedm_diffpen}
\end{figure*}

\subsection{Preliminaries on Diffusion Models}


Diffusion models are a class of generative models that synthesize data by learning to reverse a fixed noise process through iterative denoising steps.
Denoising Diffusion Probabilistic Models (DDPM)~\cite{sohl2015deep}, a widely used instantiation of this idea, define a latent variable model $p(x) = \int p(x,z) \, dz$, where $z = \{z_1, z_2, \ldots, z_T\}$ is a latent Markov chain over $T$ timesteps. The forward process $q(z|x)$ gradually corrupts data with Gaussian noise in a variance-preserving manner, and the reverse process is learned to approximate $p(x|z)$.
This formulation resembles a hierarchical Variational Autoencoder~\cite{luo2022understanding}. 
In our case, we adopt the DDIM formulation~\cite{song2020denoising}, a deterministic alternative to DDPM, while keeping the same core noise prediction structure.

In the Styled HTG setting, the conditioning $c$ typically consists of
the desired content $c_t$ (\eg~the target text) and the style $c_s$ (\eg~writer identity or visual features). 
Sampling is performed via ancestral denoising, starting from pure noise $z_0 \sim \mathcal{N}(0, I)$ and progressively refining it through a learned reverse process:
\begin{equation}
\resizebox{0.9\columnwidth}{!}{$
  x \sim p(x|c)
      = \int p(x|z_1)
           \, p(z_1|z_2)\,
           \cdots\,
           p(z_T|z_0)\,
           p(z_0)\,
           dz_{1:T}
$}
\end{equation}
We use the notation $p(x|c)$ to denote this generative process conditioned on the \textit{dual} $c$.
During training, the model learns to predict the noise $\epsilon_\theta$ added to a sample $x_0$ at timestep $t$, by using a noised input $x_t$ and conditions $c$. 
At inference, this prediction is used to iteratively reconstruct the image from noise.






\subsection{Motivation and Synopsis}

In HTG, modifying the conditioning inputs can introduce variation in the generated image, but often in an unstructured and entangled way. 
Since content and style representations are not fully disentangled in practice, perturbing the style component $c_s$ can inadvertently degrade the fidelity of the content $c_t$, resulting in unclear or semantically corrupted outputs.

To address this, we move beyond naive perturbation. Rather than using the perturbed condition directly, which may conflate content and style, we define a guidance mechanism that encourages the generation to be faithful to the intended $(c_t, c_s)$ condition, while actively steering it away from a negative pairing. This negative pairing is formed by corrupting one of the conditions (or both of them) and acts as a counterexample.
This builds naturally on the CFG framework, which interpolates between unconditional and conditional predictions to enhance content clarity. 
Instead of using an unconditional signal, we use a corrupted condition and isolate its influence by projecting out the component aligned with the original (positive) prediction. The remaining orthogonal direction provides a controlled, content-preserving signal that still allows for variability.

The core challenge is how to construct this orthogonal direction from a noisy negative prompt in a way that maintains meaningful structure and avoids pushing the model toward unrealistic outputs. In the next section, we define this construction formally and explain how it is integrated into the diffusion trajectory.

\subsection{Subspace Projection}

We define the positive and negative dual prompts based on the content-style condition. 
Let $r_t$ and $r_s$ denote the representations for $c_t$ and $c_s$, respectively. 
The clean pair $(r_t, r_s)$ serves as the positive prompt.

To simulate a negative pair, we generate noisy variants of the content and style representations by applying element-wise dropout and scaled Gaussian noise:
\begin{align}
\tilde{r}_s = \lambda_s \cdot \eta_s \cdot \mathcal{N}(0, I), \\
\tilde{r}_t = \lambda_t \cdot \eta_t \cdot \mathcal{N}(0, I), \\ \eta_s, \eta_t \sim \text{Bernoulli}(p), \nonumber
\end{align}
where $\lambda_s$ and $\lambda_t$ control the magnitude of the noise, and $p$ determines the sparsity of the active dimensions through dropout masks. 
This formulation enables selective corruption of latent attributes, encouraging the model to explore attribute-specific guidance paths rather than uniformly noisy directions. 
Empirically, we find that this stochastic masking mechanism yields more informative contrastive gradients.
Depending on the intended contrastive setup, one may perturb either the style or content representation independently, or both jointly, while keeping the other component fixed from the original (positive) pair.

Given the perturbed condition, we compute two noise predictions:
\begin{align}
    \epsilon_p &= \epsilon(x_t, t, r_t, r_s), \\
    \epsilon_n &= \epsilon(x_t, t, \tilde{r}_t, \tilde{r}_s).
    \label{eq:perrors}
\end{align}
While $\epsilon_n$ introduces perturbations, it likely contains both structured and unstructured deviations from $\epsilon_p$. To isolate a direction that influences generation without corrupting the core signal, we subtract the projection of $\epsilon_n$ onto $\epsilon_p$:
\begin{equation}
    \epsilon^* = \epsilon_n - \text{proj}_{\epsilon_p}(\epsilon_n).
\end{equation}
where the projection term is given by:
\begin{equation}
    \mathrm{proj}_{\epsilon_p}(\epsilon_n) = \frac{\langle \epsilon_n, \epsilon_p \rangle}{\| \epsilon_p \|^2} \cdot \epsilon_p.
\end{equation}
This orthogonal component $\epsilon^*$ captures contrastive variation while remaining disjoint from the intended generation direction. 

To ensure numerical stability and prevent degenerate behavior at large magnitudes, we clip the norm of $\epsilon_n$, before the projection step, using a threshold $\tau$:
\begin{equation}
    \epsilon_n \leftarrow \min\left(1, \frac{\tau}{\| \epsilon_n \|}\right) \cdot \epsilon_n.
\end{equation}
The final denoising prediction becomes:
\begin{equation}\label{eq:residual}
    \hat{\epsilon} = \epsilon_p + g(t) \cdot (\epsilon_p - \epsilon^*),
\end{equation}
where $g(t)$ is the time-dependent guidance scale, described in Section~3.4.

\subsection{Guidance Scale with Scheduling}

 The denoising behavior naturally follows a coarse-to-fine progression. 
The first timesteps of the denoising process are dominated by noise and lack a clear structure, making it undesirable to enforce a strong guidance signal and have a very heavy influence on the decisions. 
Conversely, in later ``cleaner" timesteps, the sample is refined with intricate details, and over-conditioning may hinder the preservation of subtle features and cause the appearance of artifacts. 
To address this, we influence the guidance signal across timesteps by using a triangular schedule.
For every given timestep \(t \in [0,T]\), a threshold $u_T$ is defined such that:
\begin{equation}
\gamma(t) =
\begin{cases}
\displaystyle \frac{t}{u_T}, & \text{if } t \leq u_T, \\[1ex]
\displaystyle 1 - \frac{t - u_T}{T - u_T}, & \text{if } t > u_T.
\end{cases}
\end{equation}
Here, $u_t$ controls the location of the peak in the triangle.
The overall guidance scale at timestep \( t \) is then given by multiplying $\gamma(t)$ by a base guidance factor $gs$:
\begin{equation}
g(t) = gs \cdot \gamma(t).
\end{equation}

\subsection{Sampling with Dual Orthogonal Guidance}

The proposed guidance is applied during each denoising step of the reverse diffusion process. At step $t$, we compute the standard conditional prediction $\epsilon_p$ using the given content-style pair $(c_t, c_s)$, and the negative prediction $\epsilon_n$ using the perturbed counterpart (either $\tilde{c}_s$, $\tilde{c}_t$, or both). The orthogonal direction $\epsilon^*$ is derived by subtracting the projection of $\epsilon_n$ onto $\epsilon_p$, as discussed previously.

The final residual from~\cref{eq:residual} 
is used within the DDIM sampling rule to update the sample $x_t$:
\begin{equation}
    x_{t-1} = \text{DDIMStep}(x_t, \hat{\epsilon}, t),
\end{equation}
where $\text{DDIMStep}$ denotes the deterministic transition rule at timestep $t$. This operation can be implemented directly using any scheduler that supports DDIM-style updates.

By leveraging both a faithful prompt and a structured negative variant, the proposed guidance effectively encourages content-clarity and controlled stylistic variation. The orthogonal decomposition ensures that generation is nudged along contrastive directions that preserve semantic fidelity.

\section{Experiments}
\label{sec:experiments}

\begin{figure*}[ht]
  \centering
  \includegraphics[width=\textwidth]{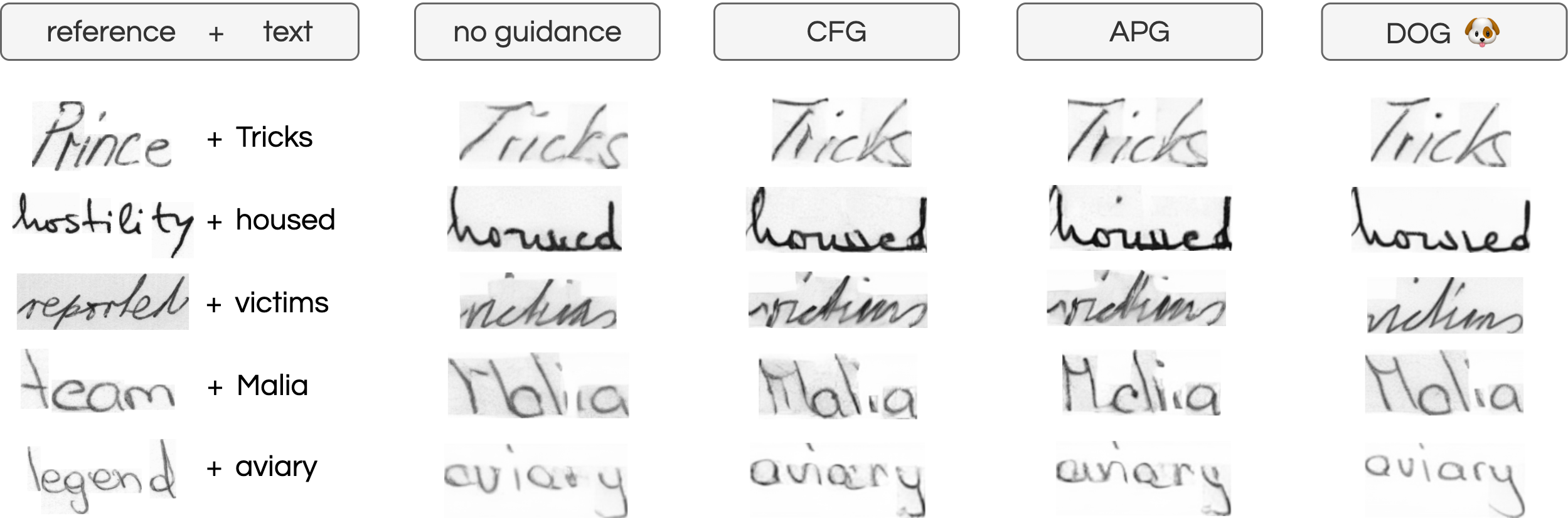}
  \caption{Qualitative comparison between guidance strategies applied to DiffusionPen when generating a target text in different styles.}
  \label{fig:comparison}
\end{figure*}





In this section, we evaluate our proposed guidance strategy, DOG, by applying it to pre-trained, off-the-shelf diffusion-based HTG models. 
We present both \textit{qualitative} and \textit{quantitative} results that demonstrate the effectiveness of DOG in improving generation quality. 
In addition, we provide ablation studies analyzing the impact of key components and hyperparameters of our method.

\subsection{Experimental Setup}

We conduct experiments with the two main existing pre-trained diffusion-based HTG backbones, DiffusionPen~\cite{nikolaidou2024diffusionpen} and One-DM~\cite{dai2025one}, trained on the IAM offline handwriting database~\cite{marti2002iam}.
We also use a version of DiffusionPen~\cite{nikolaidou2024diffusionpen} that is pre-trained on the GNHK dataset~\cite{lee2021gnhk}. 
DiffusionPen is a latent-diffusion HTG that deploys a hybrid metric- and classification-style encoder to embed style features in a few-shot setting, while One-DM operates on pixel space in a one-shot style encoding setting.

For comparison with the guidance literature, we further set up CFG~\cite{ho2021classifierfree} and APG~\cite{sadat2024eliminating} on DiffusionPen.
To this end, we re-train DiffusioPen~\cite{nikolaidou2024diffusionpen} with style and content conditions dropped with a probability of 0.2 in order to use the unconditional components necessary for CFG and APG that were not included in the original training.
We keep the HTG models in evaluation mode and use them for sampling in their default settings, integrating ours and the competitor guidance strategies as described in~\cref{sec:method}.
Hence, except for the adaptation of DiffusionPen for comparative reasons, no training process is included.
The hyperparameters for DOG are set as follows.
For the triangular scheduling, we use a peak threshold timestep of \(u_T = 700\).
Throughout all experiments, we fix the noise magnitudes to \(\lambda_s = \lambda_t = 1000\) and use a keep-probability of $p=0.75$ for dropout masks.

\begin{figure}[ht]
  \centering
  \includegraphics[width=\columnwidth]{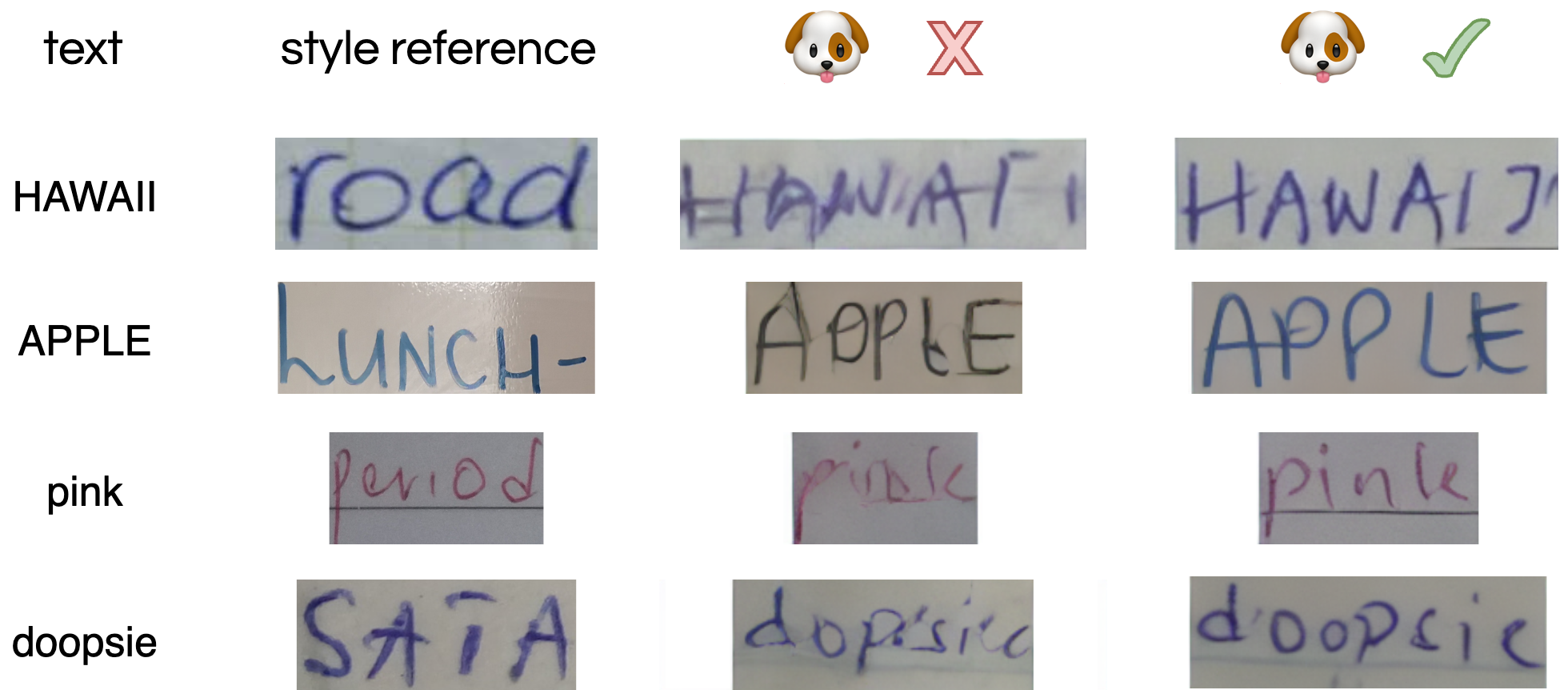}
  \caption{Qualitative examples of the proposed DOG on DiffusionPen when replicating unseen styles from the GNHK dataset.}
  \label{fig:style_gnhk_iam}
\end{figure}

\subsection{Qualitative Results}

We present qualitative examples in different scenarios showcasing the effect of our proposed method, focusing on aspects such as content robustness, style replication, and variability, while comparing with other guidance strategies.
%
\vspace{-20pt}
\paragraph{Content and Style Robustness.}
DOG appears to stabilize the generation and produce more accurate text in both seen and unseen style cases.
~\cref{fig:oov_onedm_diffpen} showcases the effect of DOG on both DiffusionPen~\cite{nikolaidou2024diffusionpen} and One-DM~\cite{dai2025one} 
when generating Out-of-Vocabulary (OOV) words using seen writer styles of the IAM database. 
Furthermore, ~\cref{fig:teaser,fig:style_gnhk_iam} show how the adaptation of DOG on DiffusionPen for the GNHK dataset~\cite{lee2021gnhk} fixes hard cases of unseen styles while preserving the intended style. 
It is clearly observed that DOG enhances the quality of the generation by consistently improving the content of the generated words while keeping the style characteristics close to the initial generation. 
Notably, DOG is model-agnostic, yielding similar content improvements when applied to both DiffusionPen~\cite{nikolaidou2024diffusionpen} and One-DM~\cite{dai2025one}.

\begin{figure}[ht]
  \centering
  \includegraphics[width=\columnwidth]{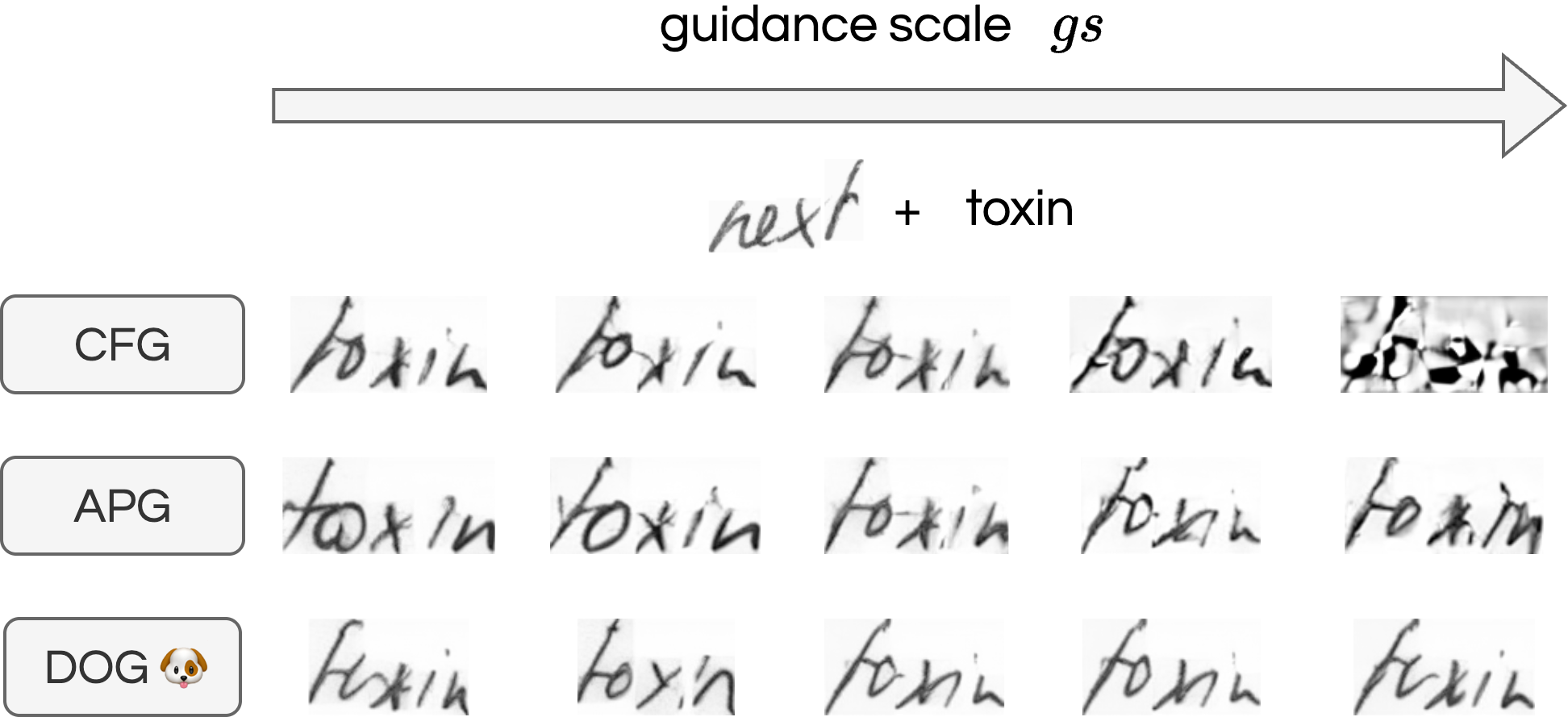}
  \caption{Comparison between different guidance strategies as guidance scale $gs$ increases. We showcase $gs$ values of 2, 5, 10, 20, and 30 (left to right column).}
  \label{fig:comparison_gs}
\end{figure}

To properly assess our method, we also compare it against the alternative guidance strategies CFG~\cite{ho2021classifierfree} and the recently introduced APG~\cite{sadat2024eliminating}. 
\cref{fig:comparison} shows comparative qualitative examples of DiffusionPen in its original form, \ie~without any guidance, and with the additional guidance strategies.
It is clear that all guidance strategies assist the content preservation.
However, our proposed method is able to generate ``cleaner'' images with less noisy artifacts.
In addition to preserving content and style, our method supports substantially higher guidance scale values, as shown in~\cref{fig:comparison_gs}. 
This robustness is enabled by the proposed scheduling strategy, which allows for more effective guidance, even in challenging cases, while reducing sensitivity to the hyperparameter $gs$.
We present further proof of this property in~\cref{subsec:quantitative}.

Finally, while the existing diffusion HTG models have the ability to generalize to datasets like IAM~\cite{marti2002iam}, in harder cases, such as GNHK~\cite{lee2021gnhk}, an unseen style might be hard to reproduce with faithful content.
In~\cref{fig:style_gnhk_iam}, we observe that using DOG, the generation process manages to approach harder cases of unseen styles during training, while preserving the content.

\vspace{-20pt}
\paragraph{Variability.}

\begin{figure}[ht]
  \centering
  \includegraphics[width=\columnwidth]{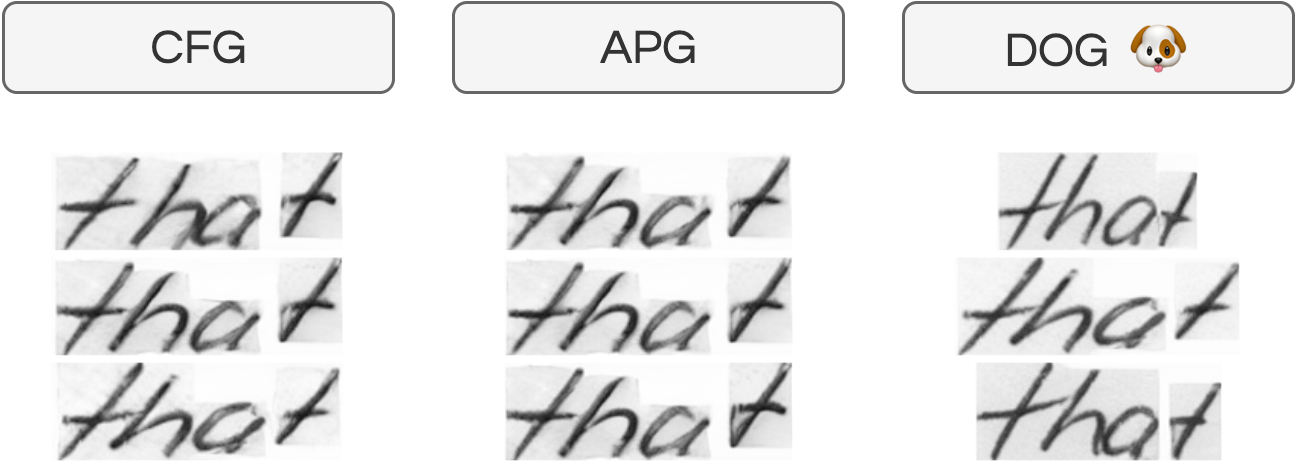}
  \caption{Comparison between CFG, APG, and the proposed DOG in terms of variance obtained from multiple samplings with different noise initializations.}
  \label{fig:variance}
\end{figure}

Diffusion-based HTG models tend to memorize the training set and often struggle to generate variations of the same words written by a specific writer, especially when only a single example of this instance exists. 
This means that the model has not learned to produce different instances of a given query word in a specific writer’s style.
This issue becomes evident when comparing DOG’s variation capabilities with CFG and APG in~\cref{fig:variance}. 
DOG consistently generates diverse instances of the same word across multiple runs, whereas CFG and APG exhibit only limited visual variation, often producing nearly identical outputs, even from random noise initialization, where different sampling output is expected.


\subsection{Quantitative Results}
\label{subsec:quantitative}

We evaluate the generated data by including Handwriting Text Recognition (HTR) experiments using the system proposed in~\cite{retsinas2022best}, in similar ways as~\cite{nikolaidou2024rethinking}.
We mainly focus on an OOV-generated set of seen writer styles from IAM to quantify the ability to produce variable styles and preserve text content.
Moreover, we compute commonly used scores such as HWD~\cite{pippi2023hwd} and FID~\cite{dowson1982frechet} on generated IAM test sets using the guidance strategies.
For this section, we proceed with DiffusionPen as a backbone, as it performs faster generation than One-DM due to its latent space operation.

\begin{table}[ht]
  \centering
  \caption{HTG\textsubscript{OOV} (CER ↓) for DiffusionPen with various guidance types and strengths. The lower the better. Bold refers to the best score per $gs$ value and underlined per guidance strategy.}
  \label{tab:oov_table}
  \renewcommand{\arraystretch}{1.1}
  \setlength{\tabcolsep}{6pt}
  \begin{tabular}{lcccc}
    \toprule
     & \multicolumn{4}{c}{Guidance} \\                         
    \cmidrule(lr){2-5}
    $g_s$ & none & CFG & APG & DOG (ours) \\                   
    \midrule
     -- & 22.8 & --    & --    & --    \\
      2 & --    & 20.1 & \underline{20.8} & \textbf{18.3} \\
     10 & --    & \underline{19.9} & \underline{20.8} & \textbf{\underline{18.1}} \\
     20 & --    & 39.4 & 24.0 & \textbf{18.4} \\
     30 & --    & 90.8 & 29.4 & \textbf{18.5} \\
    \bottomrule
  \end{tabular}
\end{table}

\vspace{-5pt}
\paragraph{OOV Words.}
We showcase how DOG improves content accuracy by creating a small set of $\sim18.6$K OOV words generated with random seen writer styles from IAM.
We quantify the effect of CFG, APG, and DOG by presenting HTR\textsubscript{OOV} for the best guidance scale $gs$ for each strategy in~\cref{tab:oov_table}.
HTG\textsubscript{OOV} refers to using an HTR trained on the real IAM corpus and testing the recognition of the OOV set, which is a Character Error Rate (CER), hence evaluating the readability.
We can see that DOG achieves better performance in terms of content accuracy with the lowest CER in the HTG\textsubscript{OOV}.

\vspace{-6pt}
\paragraph{Guidance Scale Range.}
Our proposed DOG enables larger values of the base guidance scale $gs$ as shown in~\cref{fig:comparison_gs}.
To quantify that effect and obtain the best scores for every guidance strategy, we present HTG\textsubscript{OOV} in various guidance scales in~\cref{tab:oov_table}, spanning $gs$ values from 2 to 30.
It is clear that our method gives more robust results as $gs$ increases, while CFG and APG output much more noise and artifacts, harming the readability.

\vspace{-5pt}
\paragraph{Handwritten Text Recognition (HTR).}

To ensure that the generation using our proposed guidance does not simplify the text, resulting in the high recognition performance obtained in the HTG\textsubscript{OOV} results, we incorporate generated data into the training process of an HTR system~\cite{retsinas2022best}.
To this end, we generate a large corpus of $\sim$376K samples from IAM training writer styles using DiffusionPen with and without the DOG guidance.  
We incorporate the large generated sets in the training process of the HTR along with the real data, aiming to boost the performance, and present the CER on the real validation and test sets of the IAM database. 
The results are presented in~\cref{tab:large_oov_real_table}. \begin{table}[t]
\centering
\renewcommand{\arraystretch}{1.3}
\addtolength{\tabcolsep}{6pt}
\caption{HTR performance on the real IAM validation and test sets when incorporating large-scale generated data with and without DOG. For DOG, we use $gs=20$.}
\scalebox{1}{
\begin{tabular}{lcc}
\toprule
\multirow{2}{*}{Training} & \multicolumn{2}{c}{CER $\downarrow$}\\
\cmidrule(lr){2-3}
 & validation & test\\
\midrule
Real IAM & 3.58 & 4.92\\
\midrule
DiffPen~\cite{nikolaidou2024diffusionpen} & 2.43 & 4.17\\
+DOG (ours) & \textbf{2.27} & \textbf{3.99}\\
\bottomrule
\end{tabular}}
\label{tab:large_oov_real_table}
\end{table}
In both generated cases, the performance improves; however, the generated data produced using our guidance strategy improves the recognition even more.
We should note that to avoid harming the HTR learning with too noisy data, we filter the generated data as proposed in~\cite{nikolaidou2024rethinking}.
This means that with our proposed DOG, in one generation pass, we are able to keep more data that is useful to train an HTR system.

\begin{table}[ht]
\centering
\renewcommand{\arraystretch}{1}
\setlength{\tabcolsep}{9pt}
\caption{HWD, FID, and CER scores of the IAM test set generated by DiffusionPen using no strategy or CFG, APG, and our proposed DOG. For all scores, the lower the better. Bold refers to the best result per score, and underlined refers to the best value of each score per guidance strategy.}
\label{tab:scores_results_test}
\scalebox{0.95}{
\begin{tabular}{lcccc}
\toprule
Guidance & $g_s$ & HWD\textdownarrow & FID\textdownarrow & CER\textdownarrow \\
\midrule
None & -- & 1.57 & \textbf{12.05} & 10.2 \\
\midrule
\multirow{4}{*}{CFG} 
 & 2  & 1.56 & \underline{12.39} & \underline{8.7} \\
 & 10 & \textbf{\underline{1.55}} & 13.95 & 9.0 \\
 & 20 & 1.61 & 22.16 & 22.3 \\
 & 30 & 2.40 & 134.81 & 82.1 \\
\midrule
\multirow{4}{*}{APG} 
 & 2  & 1.57 & \underline{12.79} & \underline{8.9} \\
 & 10 & \textbf{\underline{1.55}} & 13.57 & 9.7 \\
 & 20 & 1.56 & 16.63 & 11.6 \\
 & 30 & 1.62 & 21.48 & 16.4 \\
\midrule
\multirow{4}{*}{DOG (ours)} 
 & 2  & 1.65 & 22.16 & 7.8 \\
 & 10 & 1.65 & 22.16 & \textbf{\underline{7.2}} \\
 & 20 & 1.65 & 22.49 & 7.5 \\
 & 30 & \underline{1.64} & \underline{20.66} & 7.5 \\
\bottomrule
\end{tabular}}
\end{table}

\paragraph{Generation Scores.}

\begin{figure*}[ht]
  \centering
  \includegraphics[width=\textwidth]{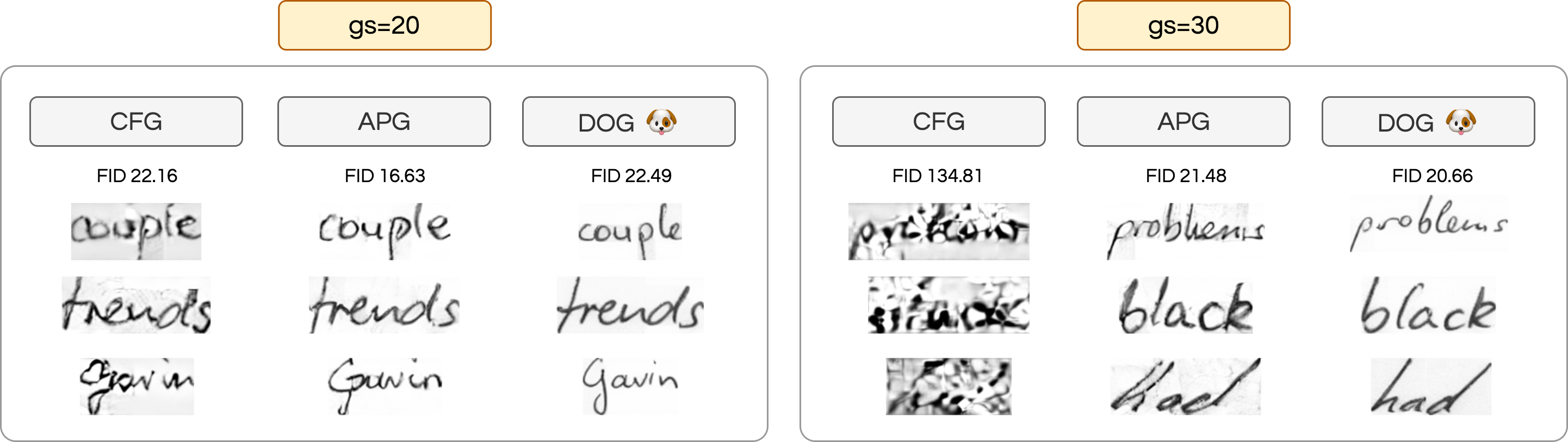}
  \caption{Qualitative results of CFG, APG, and our proposed DOG for $gs=20$ (left) and $gs=30$ (right) in correlation with FID score.}
  \label{fig:fid}
\end{figure*}
We present HWD~\cite{pippi2023hwd}, FID~\cite{dowson1982frechet}, and CER scores, comparing the IAM test set generation quality of DiffusionPen, with and without the guidance strategies, in~\cref{tab:scores_results_test}.
While our method clearly improves readability according to the CER results, HWD is slightly harmed, which is expected as the style output might drift due to the guidance.
FID is increasing by a value of 10 across all $gs$ values for our proposed DOG compared to CFG and APG, which have increased scores for $gs>20$.
However, if we look qualitatively at the outputs, it is clear that FID is not appropriate to evaluate the quality of the generation as shown in~\cref{fig:fid}.
There, we can see that for lower values of FID for $gs=20$, both CFG and APG output noisy samples, while DOG, which has the worst (highest) FID score, presents the most stable samples.
In the case of $gs=30$, we can observe that CFG outputs completely erroneous samples, which justifies the exploded FID score it obtains, while APG has a worse score, still, though, lower than the $gs=20$ of DOG that produces the most ``clean'' results.
This confirms that FID is not an appropriate score to measure HTG quality.



\subsection{Ablation}

\begin{figure}[ht]
  \centering
  \includegraphics[width=\columnwidth]
  {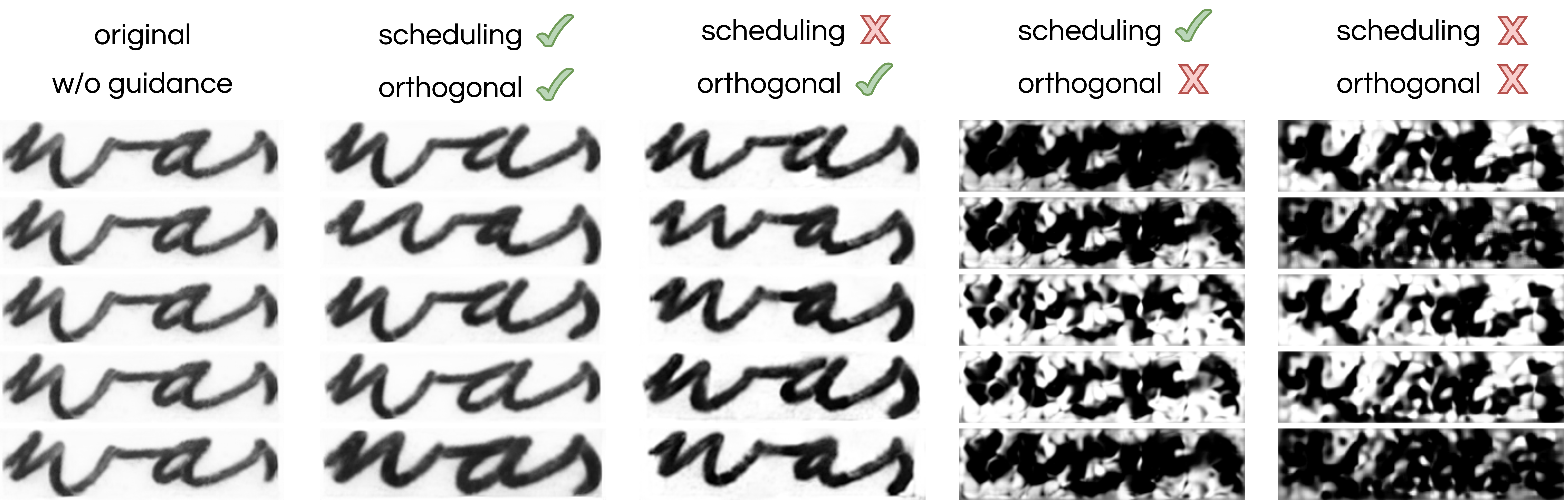}
  \caption{Ablation on the use of our proposed guidance scheduling and orthogonal component. The second column represents the full DOG method. The samples are generated using $gs=2$.}
  \label{fig:ablation_orthogonal_scheduling}
\end{figure}

We perform ablation studies on the key elements of our proposed method: the orthogonal projection and the scheduling strategy.
The importance of both components is evident as shown in~\cref{fig:ablation_orthogonal_scheduling}. 
Without the orthogonal projection, the generation collapses under the influence of high-magnitude noise, which overwhelms the meaningful style and content signals. 
Similarly, removing the scheduling component leads to noticeably noisier outputs.
In this case, we have used a value of $\lambda_s$ and $\lambda_t$ equal to $100$ to better demonstrate the effect of the examined components.
This issue becomes increasingly severe at higher $gs$ values, even when orthogonal projection is applied.

\vspace{-4pt}
\begin{figure}[ht]
  \centering
  \includegraphics[width=\columnwidth]{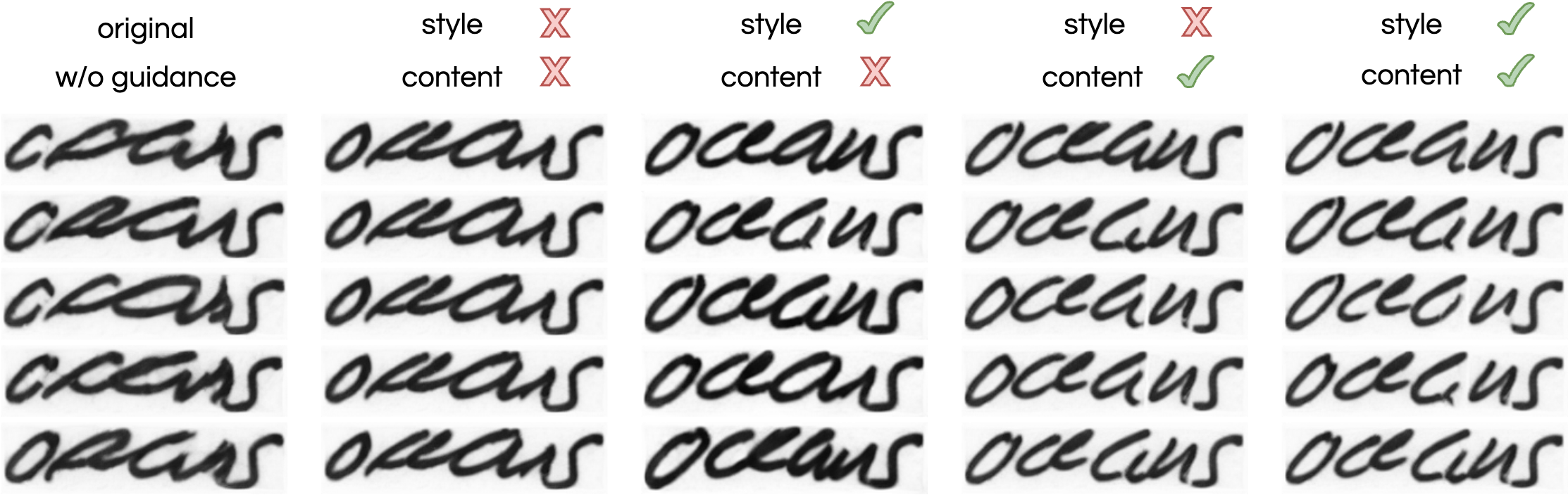}
  \caption{Ablation on the possible negative conditions (style and content) to obtain the negative direction using $gs=25$.
  Second column corresponds to using the unconditional prediction as the negative prompt, while the third and fourth column keep the content and the style, respectively, the same as the positive prompt.}
  \label{fig:ablation_style_content}
\end{figure}


Moreover, we examine the effect of using different negative conditions, namely style and content, when constructing the negative direction in~\cref{eq:perrors}, as shown in~\cref{fig:ablation_style_content}. We compare several configurations: using only the negative style or negative content while keeping the other fixed, and replacing the negative condition entirely with the unconditional prediction (second column), keeping the projection and scheduling components constant across all cases. 
The results lead to two key observations. 
First, while using the unconditional prediction improves over the original system without guidance, it provides less variability and clarity compared to using negative prompts, even when only one condition is altered. 
Second, applying a negative condition to either the style or content alone appears to be effective, supporting the flexibility of the proposed method depending on the target needs of the generation.

\begin{figure}[ht]
  \centering
  \includegraphics[width=\columnwidth]
{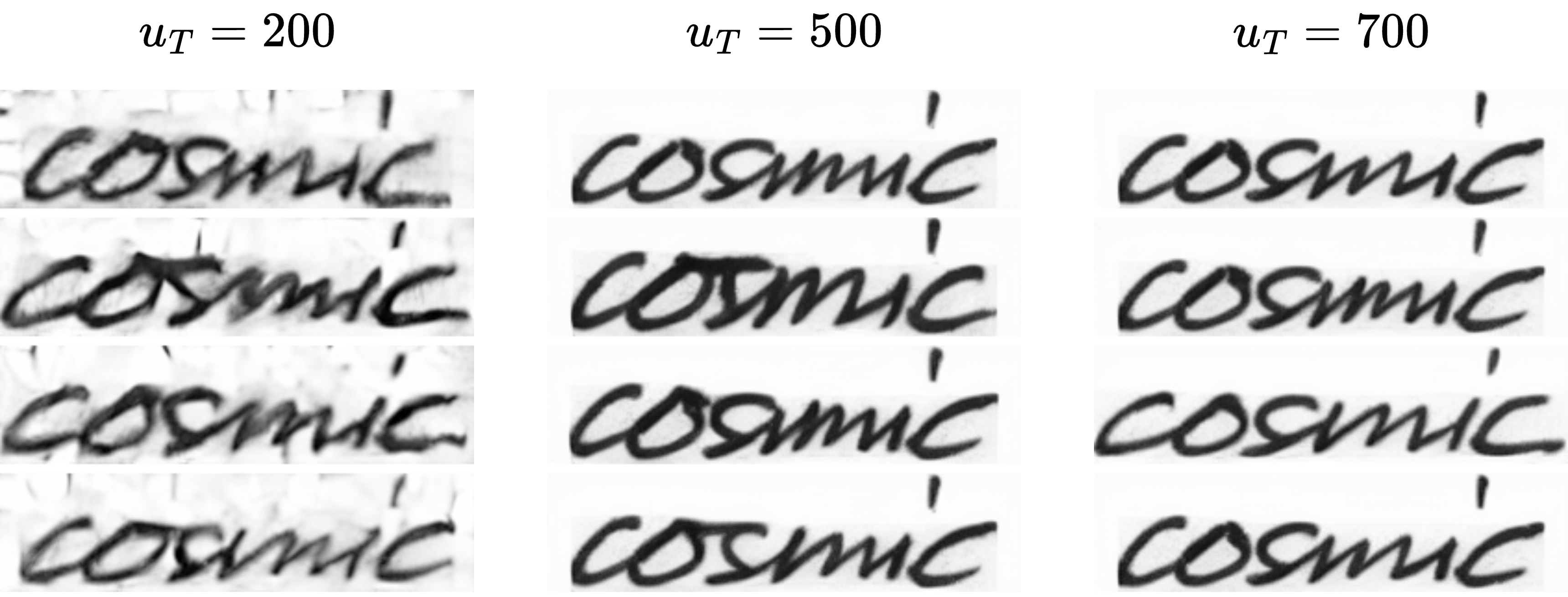}
  \caption{Ablation on the scheduling peak timestep $u_T$ showing an early (left), middle (middle), and later (right) timestep. The samples are generated using $gs=30$.}
  \label{fig:uT_ablation}
\vspace{-4pt}
\end{figure}

Finally, we experiment with early, middle, and late timestep values of the scheduling peak $u_T$ presented in the triangular scheduling of our proposed DOG.
As we can see in~\cref{fig:uT_ablation}, the early peak value choice of 200 generates noisy outputs.
A possible reason is that in later (cleaner) steps, the sample has already been formed, hence a high guidance at that point might create erosions.
Looking at the results, the earlier (noisier) timestep peak we choose for the scheduling, the more stable results we have, as the guidance is provided in a step where formation choices are still made, and hence, our choice of $u_T=700$.

\section{Conclusion and Future Work}
\label{sec:conclusion}

We introduced a simple yet effective sampling strategy for diffusion-based HTG systems named DOG.
Instead of the unconditional prediction used in traditional CFG, DOG employs a \emph{structured negative prompt} and derives an
orthogonal update that is disentangled from the positive direction, providing a dual-component condition.
We couple the guidance with a triangular, time-aware schedule and a scaling component for further robustness.
DOG can be plugged into any off-the-shelf diffusion-based HTG model without further training and improves the generation with more faithful content while preserving style variability.
We compare DOG with existing CFG and APG sampling strategies, showcasing superior and more stable generation results.
Through a combination of qualitative and quantitative results, we also show how problematic the use of FID is in domains outside of natural images, like HTG.
While our work proves robust in batch sampling, there is still room for improvement, as each sample may benefit from its own ideal guidance scale.
In general, DOG can serve as a preliminary exploration for future research on guiding the generation of handwritten words with more robust results, enhancing content accuracy while preserving the style.


    


\clearpage

\section*{Acknowledgment}
\noindent
The computations and data handling were enabled by the Berzelius resource provided by the Knut and Alice Wallenberg Foundation at the National Supercomputer Centre at Linköping University.

{
    \small
    \bibliographystyle{ieeenat_fullname}
    \bibliography{main}
}


\end{document}